\documentclass[10pt, a4paper]{article}
\usepackage{lrec2022} 
\usepackage{multibib}
\newcites{languageresource}{Language Resources}

\usepackage{amsmath}
\DeclareMathOperator*{\argmax}{arg\,max}
\usepackage{tikz}

\usepackage[caption=false]{subfig}
\usepackage{booktabs}
\usepackage{float}
\usepackage{adjustbox}
\usepackage{pifont}
 \usepackage{array,multirow,graphicx}

\usepackage{amssymb}
\usepackage{algorithm}
\usepackage[noend]{algpseudocode}
\usepackage{booktabs}
\usepackage{multirow}

\usepackage{tabu}
\usepackage{array}
\usepackage{graphicx}
\usepackage{tabularx}
\usepackage{mwe}

\usepackage{float}
\usepackage{caption}
\usepackage{xcolor}
\usepackage{tikz}
\usepackage{subfig}
\usetikzlibrary{decorations.pathreplacing}

\usepackage{titlesec}
\titleformat{\section}{\normalfont\large\bfseries\center}{\thesection.}{1em}{}
\titleformat{\subsection}{\normalfont\SmallTitleFont\bfseries\raggedright}{\thesubsection.}{1em}{}
\titleformat{\subsubsection}{\normalfont\normalsize\bfseries\raggedright}{\thesubsubsection.}{1em}{}
\renewcommand\thesection{\arabic{section}}
\renewcommand\thesubsection{\thesection.\arabic{subsection}}
\renewcommand\thesubsubsection{\thesubsection.\arabic{subsubsection}}

\usepackage{epstopdf}
\usepackage[utf8]{inputenc}

\usepackage{hyperref}
\usepackage{xstring}

\usepackage{color}

\title{Context-based Virtual Adversarial Training \\ for Text Classification with Noisy Labels}

\name{Do-Myoung Lee$^{1*}$, Yeachan Kim$^{2*}$, Chang-gyun Seo$^{3}$} 

\address{$^{1}$ShinhanCard, $^{2}$Deargen Inc., $^{3}$GC Company \\
         $^{1}$domyoung89@shinhan.com, $^{2}$yeachan@deargen.me, $^{3}$ocean@gccompany.co.kr}

\abstract{
Deep neural networks (DNNs) have a high capacity to completely memorize noisy labels given sufficient training time, and its memorization unfortunately leads to performance degradation. Recently, virtual adversarial training (VAT) attracts attention as it could further improve the generalization of DNNs in semi-supervised learning. The driving force behind VAT is to prevent the models from overffiting to data points by enforcing consistency between the inputs and the perturbed inputs. These strategy could be helpful in learning from noisy labels if it prevents neural models from learning noisy samples while encouraging the models to generalize clean samples. In this paper, we propose context-based virtual adversarial training (ConVAT) to prevent a text classifier from overfitting to noisy labels. Unlike the previous works, the proposed method performs the adversarial training in the context level rather than the inputs. It makes the classifier not only learn its label but also its contextual neighbors, which alleviate the learning from noisy labels by preserving contextual semantics on each data point. We conduct extensive experiments on four text classification datasets with two types of label noises. Comprehensive experimental results clearly show that the proposed method works quite well even with extremely noisy settings. 
 \\ \newline  }

\begin{document}

\maketitleabstract

\section{Introduction}
\def\thefootnote{*}\footnotetext{Equal contribution.}\def\thefootnote{\arabic{footnote}}
Deep neural networks (DNNs) have shown human-level performance in various domains, such as image classification, machine translation, and speech recognition. To achieve such an ability, it is indisputably evident that we have to collect a large amount of training data. As labeling such data is laborious and expensive, previous works utilize search engine \cite{blum2003noise,li2017webvision} or crowdsourcing \cite{yan2014learning,yu2018learning} to collect labeled dataset. Unfortunately, these low-cost approaches introduce low-quality annotations with \textit{label noise}. It causes DNNs to completely memorize such label noises (i.e., nearly 100\% training accuracy) \cite{zhang2016understanding}, deteriorating generalization capability  \cite{frenay2013classification,sukhbaatar2014training}.

To deal with label noises, recent works primarily propose loss correction approaches. These methods directly correct loss function (e.g., cross-entropy, mean squared error) or the probabilities used to compute it. For example, \cite{zhang2016understanding} learns sample weighting scheme via an auxiliary network (called \textit{MentorNet}) and applies it to noisy labels such that corrupt data could get nearly zero sample weights on the loss function. On the one hand, \cite{han2018co} utilizes an intervention between two same networks with small-loss samples which are considered as clean. However, there exists only a single work to handle label noise on natural language. \cite{jindal2019effective} utilize a noise transition matrix for text classification. This method is useful at text classification on label noises, but it turned out that estimating real noise transition matrix is difficult \cite{jiang2018mentornet,han2018co} especially when the number of classes is large. 

Recently, adversarial training  at attracts attention as it could prevent networks from misclassifying an image that contains a small perturbation (e.g., adversarial examples). The at enforces the classifier to make consistent predictions on synthetic inputs that have small, approximately worst-case perturbations. \cite{miyato2015distributional} extends the idea of at to the semi-supervised regime by removing label dependency on perturbations, which is called a virtual adversarial training (VAT). Adversarial training indirectly have a label smoothing effect as it prevents models from predicting given labels with high confidence by anisotropically smoothing around each data point. Several recent works have shown that label smoothing is effective as a means of coping with label noise \cite{lukasik2020does} and preventing memorization \cite{xie2016disturblabel}. Based on such observations, we further study whether the adversarial training methods, which is an indirect label smoothing method, are useful at training a robust classifier on label noise.

In this paper, we propose a \textit{context-based virtual adversarial training} (ConVAT) to build a robust text classifier on label noise. We inherently follow a fundamental strategy of VAT. 
Unlike the previous work \cite{miayto2016virtual}, ConVAT performs the adversarial training on the context-level feature space not the word-level. To that end, we solve min-max optimization by following two steps: \textit{formulating perturbation} and \textit{smoothing}. In the first phase, we calculate the worst-case perturbation into an adversarial direction that could maximize the classification loss on the given samples. We then minimize the distributional distance between a normal sample and a perturbed sample to learn robust classifier on adversarial perturbations. This strategy allows us to train a label noise-robust classifier without placing a burden in a network computation.

In order to show the strength of the proposed method, we conduct extensive experiments on four different datasets with the different kinds of label noise. Comprehensive evaluation results clearly show that ConVAT outperforms the state-of-the-art method in text classification with noisy labels. The in-depth analysis demonstrates that the proposed method have a strong advantage over previous adversarial methods in terms of time and memory complexity. Our code and dataset are publicly available$^{1}$.
\def\thefootnote{1}\footnotetext{\url{https://github.com/domyounglee/baseline/tree/convat}}\def\thefootnote{\arabic{footnote}}

\begin{figure}
    \centering
    \subfloat[]{\includegraphics[width=0.25\textwidth]{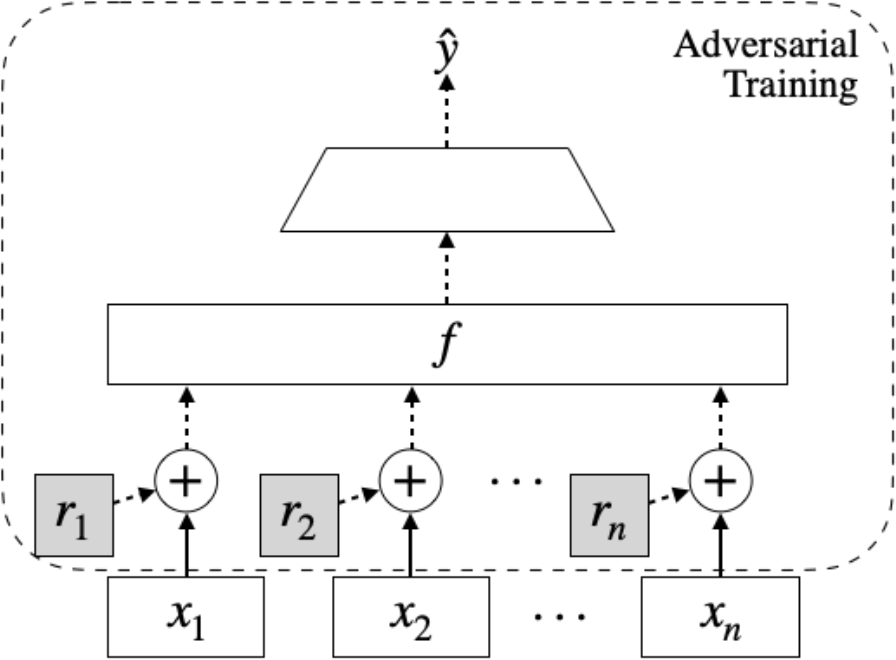}} 
    \subfloat[]{\includegraphics[width=0.22\textwidth]{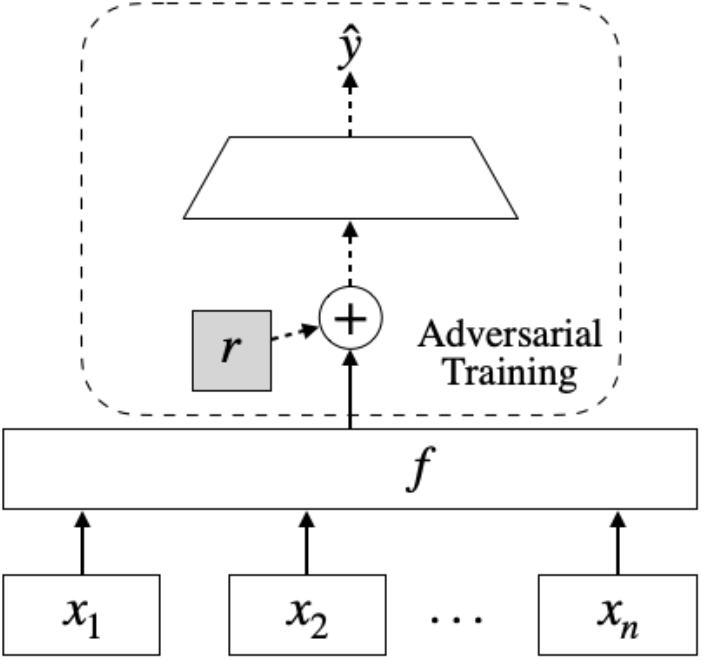}} 
    \caption{ Overall training procedure in (a) virtual adversarial training (b) context-based virtual adversarial training (ConVAT). Dotted line indicates a duplicated propagation path to generate adversarial perturbation.}
    \label{fig:model}
\end{figure}

\section{Related Works}
\subsection{Adversarial Training}

Adversarial examples are synthesized input that is created by making small perturbations to the input, which are designed to significantly increase the loss incurred by a machine learning model\cite{szegedy2013intriguing,goodfellow2014explaining}. Adversarial training \cite{goodfellow2014explaining} and virtual adversarial training \cite{miyato2015distributional} are proposed to build a robust classifier against adversarial examples. The difference between the two works lies in how to make an adversarial perturbation. The former takes loss values from label information and makes a worst-case perturbation which prevents the classifier from correctly classifying an input image into a given label. Rather than using label information, the latter only uses the output distribution of around data points since nearby points should be similar in terms of their labels \cite{han2018co}. Recently, VAT attracts attention from various domains. For example, \cite{deng2019batch} applies this idea to graph convolutional neural networks for node classification by generating perturbations to all nodes in networks. \cite{chen2020seqvat} applies VAT into sequence labeling tasks by combining VAT with the conditional random field.


\subsection{Noisy Label Learning}

Several works try to train a robust classifier on label noise. Recently, these works popularly adopt loss correction approaches which directly correct loss function or the probabilities used to compute it. One common approach is modeling the noise transition matrix which defines the probability of one class flipped to another one \cite{natarajan2013learning,patrini2017making}. However, it turns out that estimating real noise distribution is difficult \cite{jiang2018mentornet,han2018co}. As a different line of work, several works utilize inference modules to differentiate clean and noisy samples using a neural network \cite{jiang2018mentornet,lee2018cleannet}, a graphical model \cite{Xiao_2015_CVPR,vahdat2017toward}. These works are quite useful in learning a robust classifier, but they require the extra clean data or expensive noise detection, which is unpractical in real-world applications. Refined training strategies is another technique to handle label noise. These works typically use two same networks, which have different learning ability, and update parameters based on their predictions such as agreement \cite{wei2020combating}, disagreement \cite{malach2017decoupling,yu2019does} between two networks. 


\section{Methodology}
In this section, we describe context-based virtual adversarial training, coined ConVAT, in details. We first formulate the problem definition for the noisy label learning (Section~\ref{sect:def}). Then, we take a contextual viewpoint for the classification (Section \ref{sect:view}). Lastly, we elaborate {ConVAT} with the contextual viewpoint (Section \ref{sect:method}). The training procedure is straightforward and described in Figure \ref{fig:model}.

\subsection{Problem Definition}\label{sect:def}
As we stated before, collecting large-scale labeled datasets with low-cost approaches (e.g., search engine, crowdsourcing) results in label noise in the datasets. In other words, the label $y$ is likely to be flipped into another class $y'$ by human annotators.

Let the training dataset be denoted by $\mathcal{D} = \{(\textbf{X}_{1}, y_1), (\textbf{X}_{2}, y_2), ..., (\textbf{X}_{N}, y_N)\}$ where $\textbf{X}_{i}$, $y_i$ is the input and the class of $i$-th sample, respectively. Each input $X_i$ is the concatenated matrix of the word vectors $x_j$ where subscript $j$ indicates the $j$-th word in the given text. The noisy dataset can be represented as follows:
\begin{multline}
    \mathcal{D}' = \{(\textbf{X}_{i},y'_i) | \textbf{X}_{i} \in{R^{T_i\times{d}}}, \\ y'_i \in \{1, 2, ..., K\}, i=1,..,N\}
\end{multline}
where $T_i$ is the $i$-th sentence length, $d$ is the dimension of a word vector, and $K$ is the number of the classes.

\begin{figure}[th]

\centering
\includegraphics[width=0.48\textwidth]{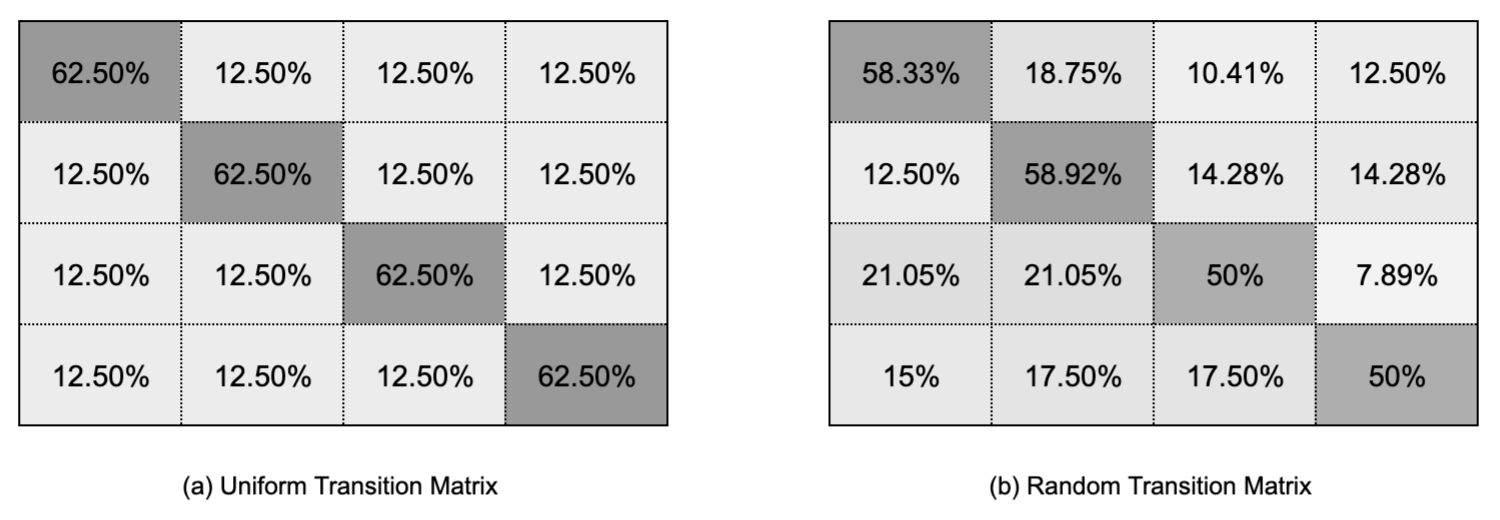}

\caption{Examples of the two types of noise transition matrix $\Phi \in R^{K\times{K}}$}
\label{noise-mat}
\end{figure}

To corrupt dataset $\mathcal{D}$ into $\mathcal{D}'$, we assume a Noise Transition Matrix  $\Phi \in R^{K\times{K}}$ filled with probabilities of flipping from one class to another. 
This follows the same settings with \cite{jindal2019effective}. We define two types of Label Noise as follows and depicted each Noise Transition Matrix in Figure \ref{noise-mat}:
\begin{itemize}
\item \textbf{Uniform label noise}: the labels are flipped from one class to another with the same probability across all the classes.

\item \textbf{Random label noise} the labels are flipped from one class to another based on a certain random distribution across all the classes.
\end{itemize}

Our goal is to train a robust text classifier from the noisy dataset $D'$. In order to achieve such objective, preventing the classifier from memorizing label noises is significant to avoid performance degradation.






\subsection{Classification from a context viewpoint}\label{sect:view}

Given a task-specific network (e.g., CNN, Transformer), we denote the term \textit{context vector} as last layer activation before the softmax layer. The \textit{context vector} contains rich contextual information which is central to various NLP tasks, such as word sense disambiguation, named entity recognition, coreference resolution\cite{melamud2016context2vec}. Formally, given a input $\textbf{X}_{i}$, a context vector $\textbf{c}_{i}$ can be written as $ \textbf{c}_{i} =  f(\textbf{X}_{i})$ where $f$ is the function of a task-specific network. We define the probability of the flipped class $y'_i$ given the context vector $\textbf{c}$ and weights  $\textbf{W}_c\in{R^{v\times{K }}} $, $v$ is the dimension of the context vector,  as follows:

\begin{align}
P( \textbf{c}_{i}, y'_i; \theta_{sm})=\frac{e^{\textbf{c}_{i}^\top \textbf{w}_{y'_i}}}{\sum_{k=1}^{K}e^{\textbf{c}_{i}^\top \textbf{w}_{k}}}
\label{eq:1}
\end{align}
where $\theta_{sm}$ is the trainable parameter $\textbf{W}_c$ of the softmax layer. We also denote the parameter of the entire network including $\theta_{sm}$ as $\theta$. 

Based on the probability function $P$, we define the loss function $l$ with cross entropy as follows :
\begin{align}
l( \textbf{c}_{i}, y'_i; \theta)=-\frac{1}{N}\sum_{i \in \textit{N}}^{}\log_{}P( \textbf{c}_{i}, y'_i; \theta)
\label{ce}
\end{align}

However, several works have shown that cross-entropy, which is a \textit{de-facto} standard loss function for the text classification, is vulnerable to label noises \cite{ghosh2017robust} due to a memorization effect. In the following subsection, we will describe how can we sidestep the memorization effect.


\subsection{Context-based Virtual Adversarial Training}\label{sect:method}




The goal of both AT and VAT is to train a robust classifier against adversarial inputs. To prevent corruption from an input, most works usually add a perturbation into an input (i.e., word embeddings) which is located in most-bottom architecture.  However, we have to prevent corruption from an output layer due to label noises. We thus mainly focus on the output layer (i.e., softmax layer), which converts a context vector into a categorical distribution, and our strategy is to minimize the distributional distance between the context vector and the perturbed context vector.
\begin{multline}
\Delta_{\mathrm{KL}}(\textbf{r}, \textbf{c}_{i},\hat{\hskip3.5mm\theta_{sm}}) \equiv {} \\
\mathrm{KL}[P( \textbf{c}_{i}, \cdot \; ; \hat{\hskip3.5mm\theta_{sm}})||\;P(\textbf{c}_{i}+\textbf{r}_{i}^{\mathrm{c-adv}}, \cdot \; ; \hat{\hskip3.5mm\theta_{sm}})] \\ \text{ where } \\
\textbf{r}_{i}^{\mathrm{c-adv}}  \equiv \argmax_{\textbf{r}}{ \Delta_{\mathrm{KL}}(\textbf{r}, \textbf{c}_{i},\hat{\hskip3.5mm\theta_{sm}}), \; ||\textbf{r}||_2\leq\epsilon }
\label{radius1}
\end{multline}
where $\textbf{r} \in R^{v}$ is a random \textit{perturbation vector} and $\mathrm{KL}[\textbf{p}||\textbf{q}]$ is a KL-divergence between distributions $\textbf{p}$ and $\textbf{q}$.
We refer to $\textbf{r}_{i}^{\mathrm{c-adv}}$ as \textit{contextual adversarial perturbation vector} which has a direction of maximum perturbation called the adversarial direction. However, calculating the exact value of $\textbf{r}_{i}^{\mathrm{c-adv}}$ is intractable. 
 With the linear approximation method of \cite{miayto2016virtual}, the resulting adversarial perturbation is redefined as follows:
\begin{multline}
\textbf{r}_{i}^{\mathrm{c-adv}} \approx \epsilon \cdot \frac{\textbf{g}}{||\textbf{g}||_2} \text{ where } \\
g = \nabla_{\textbf{r}} \mathrm{KL}[P( \textbf{c}_{i}, \;\cdot\;  ; \hat{\theta}_{sm}) ||  P( \textbf{c}_{i}+\textbf{r}, \;\cdot\;  ; \hat{\theta}_{sm})]|_{\textbf{r}=\xi \textbf{d}}
\label{radius2}
\end{multline}
where $\textbf{d} \in R^v$ is a normalized random unit vector and $\epsilon, \xi$ is a norm constant. This approximation coincides with a single iteration ($t=1$) of the power method as in previous work \cite{miayto2016virtual}. Since the goal of the above process is to calculate the perturbation, \textit{we do not update any parameters in this step}.  With a view to maintaining the uniformity of context vector, we define the context-driven label smoothing (\textbf{CLS}) :
\begin{equation}
\mathrm{\textbf{CLS}}(\textbf{c}_{i}, \theta_{sm}) \equiv \Delta_{\mathrm{KL}}(\textbf{r}_{i}^{\mathrm{c-adv}} ,\textbf{c}_{i},\theta_{sm}) 
\label{eq1}
\end{equation}
Adding \textbf{CLS} term to the loss function, the label is smoothed by tying the context vector with its neighboring perturbed vectors. The model both learns its label and its most distant neighbor. Since neighbors have useful intermediate representations even with label noises \cite{bahri2020deep}, \textbf{CLS} could make the classifier robust against label noises. Therefore, we redefine the loss function as:

\begin{multline}
l( \textbf{c}_{i}, y'_i; \theta)=\\-\frac{1}{N} \sum_{i \in \textit{N}}^{}\log_{}P( \textbf{c}_{i}, y'_i; \theta)+ \lambda \frac{1}{N}\sum_{i \in \textit{N}}^{}\mathrm{\textbf{CLS}}(\textbf{c}_{i}, \theta_{sm})
\label{final-loss}
\end{multline}
where $\lambda$ is a controlling factor of label smoothing and it adjusts the degree of how much the context vector count on its neighbors. We show our entire algorithm in Algorithm 1. 

\begin{algorithm}
\caption{\textbf{ConVAT} loss function }
\label{alg:conVAT}
\begin{algorithmic}[1]

\State \textbf{Inputs} : \\
context vector $\textbf{c} \in R^{v}$, logit vector $\textbf{z} \in R^K$ \\,  normalized random unit vector $\textbf{d} \in R^v$ 

\State \textbf{Output} : \\
KL-divergence $\Delta_{\mathrm{KL}}$ between $\textbf{z}$ and  $\hat{\textbf{z}}$

\\
$P(\mathrm{\textbf{x}})=\textbf{x}^\top \textbf{W},   \mathrm{\textbf{W}}\in{R^{v\times{K}}}$
\\
\State \textbf{Algorithm} : 
\Function {ConVAT}{$\textbf{c},\textbf{d},\textbf{z}$}

\For{  1..\textit{t} }
\State $\hat{\textbf{z}}$ = $P(\mathrm{\textbf{c}}+ \textbf{r})$
\State $\textbf{g} \gets \nabla_\textbf{r}{\Delta_{KL}(\textbf{z}, \hat{\textbf{z}})}|_{\textbf{r}=\textbf{d}}$
\EndFor
\State $\textbf{r}^{\mathrm{c-adv}}=\epsilon \cdot \textbf{g}/\left\|{\textbf{g}}\right\|^2 $
\State $\hat{\textbf{z}}$ = $P(\mathrm{\textbf{c}}+\textbf{r}^{\mathrm{c-adv}})$

\State \textbf{return} $\Delta_{KL}(\textbf{z}, \hat{\textbf{z}})$
\EndFunction {}
\end{algorithmic}
\end{algorithm}





\section{Experiments}
In this section, we conduct extensive experiments to verify the efficacy of our method compared to previous works. We describe the datasets and experimental settings in Section 4.1 and 4.2, respectively. We provide evaluation results compared to the state-of-the-art method in Section 4.3. 


\subsection{Datasets} 

In experiments, we use popular text classification datasets, which are SST-2 \cite{socher2011semi}, TREC \cite{voorhees1999trec}, Ag-News \cite{zhang2015character} and DBpedia \cite{zhang2016understanding}, to evaluate the proposed method. We explain the datasets in below:
\begin{itemize}
\item \textbf{SST-2} : Stanford Sentiment Treebank dataset for predicting the sentiment of movie reviews. The dataset is composed of positive or negative reviews.
\item \textbf{TREC}: The TREC dataset is a dataset for question classification consisting of open-domain, fact-based questions divided into broad semantic categories. We use the six-class (TREC-6) version of the dataset.
\item \textbf{Ag-News}: The AG's news topic classification dataset is constructed by choosing the 4 largest classes from the news corpus. 
\item \textbf{DBpedia}: DBpedia is a project aiming to extract structured content from the information created in Wikipedia. We used the classic 14 class version of the dataset.
\end{itemize}


We conduct experiments on two types of label noise (i.e., \textit{Uniform} and \textit{Random} noise) and evaluate the classification accuracy in the settings of different noise rates. For example, if the noise rate is 0.5, the correct class label is flipped with a 50\% probability to other labels. 





   

\begin{table*}[ht]
\begin{adjustbox}{max width=\textwidth}
\begin{tabular}{@{}c|l|c|ccccccc||ccccccc@{}}
\toprule
                 &                 & &\multicolumn{7}{c|}{{Uniform} (Symmetric)}                               & \multicolumn{7}{c}{{Random} (Asymmetric)}                     \\ \midrule \midrule
\multirow{5}{*}{{{\rotatebox[origin=c]{90}{AG-News}}}} & \textbf{Noise}  & 0.0     & 0.1     & 0.2   & 0.3    & 0.4   & 0.5    & 0.6     & 0.7    & 0.1     & 0.2    & 0.3    & 0.4   & 0.5    & 0.6    & 0.7   \\ \cmidrule(l){2-17} 
                         & CNN             & 92.31   & 89.96   & 87.42 & 84.55  & 79.96 & 75.42  & 68.78   & 59.94  & 89.71   & 86.11  & 79.05  & 76.04 & 65.09  & 45.79  & 38.12 \\
                         & {VAT}        & 92.55   &  91.4 & 90.8 & 90.03  & 89.9 & 89.21  & 88.13 & 87.74  & 92.23  & 90.93  & 90.4  & 90.0  & 87.92  & 60.93  & 52.04 \\
                         & TransMat  &\textbf{ 92.53 }  & \textbf{92.36 }  & 91.8  & 91.13  & 90.17 & 89.85 & 88.9    & 88.52  &\textbf{91.98}  & 91.5   & 90.56  & 89.75 & 83.22  & 60.22 & 49.31 \\ 
                         & ConVAT (Ours.)          & 92.44   & 92.02 &\textbf{ 91.85 }& \textbf{91.24 } & \textbf{91.01} & \textbf{90.70} & \textbf{89.85}   &\textbf{88.92}  & 91.66   & \textbf{91.52}  & \textbf{91.06}  & \textbf{89.86} & \textbf{88.28}  & \textbf{62.93}  & \textbf{53.94} \\ \midrule
\multirow{5}{*}{{{\rotatebox[origin=c]{90}{TREC}}}}    & \textbf{Noise}  & 0.0     & 0.1     & 0.2   & 0.3    & 0.4   & 0.5    & 0.6     & 0.7    & 0.1     & 0.2    & 0.3    & 0.4   & 0.5    & 0.6    & 0.7   \\ \cmidrule(l){2-17} 
                         & CNN             & 92.8    & 87.27   & 83.07 & 75     & 69.13 & 61.53  & 50.13   & 39.8   & 85.93   & 82.2   & 74     & 68.4  & 53.53  & 48.2   & 31.47 \\
                         & {VAT}        & 91.52   & 91.56   & 90.17 & 88.31  & 83.15 & 77.51  & 68.1   & 66.13  & 90      & 90.2   & 87.12  & 86.28  & 79.9   & 68.92  & 48.86 \\
                         & TransMat  & 92.6    & 91.53   & 90    & 86.1   & 84    & 82.4   & 78.67   & 74.4   & \textbf{90.78}   & \textbf{89.6}   & 85.8   & 84.5  & 74.4   & 73.4   & 48.2  \\ 
                         & ConVAT (Ours.)          &\textbf{ 92.9}    & \textbf{92.06 }  & \textbf{91.6}  & \textbf{89.58}  & \textbf{87.27} & \textbf{85.48}  & \textbf{79.47}   & \textbf{78.01}  & 90.12   & 89.53  & \textbf{87.73}  & \textbf{89.2}  & \textbf{82.93}  & \textbf{76.93}  & \textbf{49.4}  \\ \midrule
\multirow{5}{*}{{{\rotatebox[origin=c]{90}{DBpedia}}}} & \textbf{Noise}  & 0.0     & 0.3     & 0.5   & 0.7    & 0.75  & 0.8    & 0.85    & 0.9    & 0.3     & 0.5    & 0.7    & 0.75  & 0.8    & 0.85   & 0.9   \\ \cmidrule(l){2-17} 
                         & CNN             & \textbf{99.01}   & 95.19   & 89.59 & 74.01  & 67.73 & 57.87  & 47.48   & 34.01  & 94.72   & 86.08  & 62.87  & 53.13 & 40.78  & 26.6   & \textbf{12.42} \\
                         & {VAT}        & 98.04   & 97.94   & 96.81 & 96.61  & 95.52 & 94.33  & 94.13   & 93.53  & 99.01   & 94.86  & 94.66  & 93.1  & 91.67  & 50.62  & 10.27 \\
                         & TransMat  & 98.78   & 95.10  & 98.01 & 97.46 & 96.95 & 96.24 & 95.23 & 92.185 & 98.18  & 97.24 & 91.39 & 88.28 & 76.73 & 42.76  & 9.29 \\ 
                         & ConVAT (Ours.)     &     98.86 &  \textbf{98.65}& \textbf{98.39} & \textbf{97.97} &  \textbf{97.67}& \textbf{97.69} & \textbf{96.77} & \textbf{95.36} & \textbf{98.58} & \textbf{98.17} & \textbf{96.1} & \textbf{95.08} & \textbf{90.55} & \textbf{54.01} & 10.12  \\ \midrule
\multirow{5}{*}{{{\rotatebox[origin=c]{90}{SST-2}}}}   & \textbf{Noise}  & 0.0     & 0.1     & 0.2   & 0.3    & 0.4   & 0.45   & 0.47    & 0.5    & 0.1     & 0.2    & 0.3    & 0.4   & 0.45   & 0.47   & 0.5   \\ \cmidrule(l){2-17} 
                         & CNN             & \textbf{87.27}   & 83.29   & 79.08 & 73.42  & 64.03 & 58.1   & 54.73   & 49.7   & 81.44   & 75.58  & 71.88  & 63.39 & 57.12  & 55.81  & 52.32 \\
                         & {VAT}        & 87.82 & 86.35   & 84.21 & 82.89  & 79.73 & 69.76  & 63.4    & 48.74  & 85.17   & 85.35   & 79.9   & 77.8  & 69.96  & 68.56  & 46.6 \\
                         & TransMat & 87.04   & 85.19   & \textbf{85.48} & \textbf{82.59}  & 76.94 & 64.63  & 60.2    & 48.63  & \textbf{86.27}   & \textbf{84.73}  & 80.46  & 73.37 & 60.54  & 61.43  & 51.71 \\ 
                         & ConVAT (Ours.)          & 87.25   & \textbf{85.53}   & 84.23 & 82.48  & \textbf{80.83} & \textbf{68.41}  & \textbf{63.9}    & \textbf{51.18}  & 85.59   & 83.74  & \textbf{82.68}  & \textbf{78.47} & \textbf{75.94}  & \textbf{68.95}  & \textbf{53.13} \\ \bottomrule
\end{tabular}
\end{adjustbox}
\caption{Performance evaluation results on entire datasets with different kinds of noise (Uniform, Random). The best results are highlighted in \textbf{boldface}. }
\label{performance}
\end{table*}

\subsection{Experimental Settings}

In this experiment, we use single-layered convolutional neural networks \cite{kim2014convolutional} for text classification. The pre-trained GloVe \cite{pennington2014glove} is used to initialize word vectors in a vocabulary. This backbone model is denoted as \textbf{\textit{CNN}}. The comparison methods are equally applied to this model.

We mainly compare our method \textbf{ConVAT} with \cite{jindal2019effective} which is the first work that handles label noises in text classification and is denoted as \textbf{{TransMAT}}. When it comes to the regularization hyper-parameter for \textit{\textbf{TransMat}}, we choose the best performing value for each task in uniform and random, respectively. We also compare the vanilla \textbf{VAT}, which performs the adversarial training in input-level feature space, to verify the strength of the context-level adversarial training. To fairly compare with our work, we equally set the network's hyper-parameters (e.g., the number of filters, the size of the window) and these are tuned using the development set. We reproduce all the results in the same settings to compare the performance in the same noise datasets.

In our method, $\lambda$ is set to 1.0 and epsilon $\epsilon$ differently for each dataset (details in Section 5.2). The rest of the hyper-parameters is the same as \textit{\textbf{TransMat}} for a fair comparison. In training steps, we do early stopping based on development set accuracy\footnote{Labels in the development set are also flipped in our experiment.}. Entire models are implemented using Pytorch framework. We train the models on a single NVIDIA GeForce RTX 2080 Ti with 11GB of RAM.
Our code and dataset are available at \url{https://github.com/domyounglee/baseline/tree/convat}.

\subsection{Experimental Results}


Table \ref{performance} shows the overall comparison results on the four datasets with different levels and types of label noises. We test all experiments in five times and report the average test accuracy. As can be seen from the table, the performance of the backbone model {CNN} is largely decreased as the noise rate increases both in two types of noise. However, {TransMAT}, ConVAT  recover the performance loss even in extreme noise settings (i.e., 90\% noisy dataset). Between these two works, ConVAT  outperforms {TransMAT} in most settings and datasets. When the noise ratio is large, the performance gap becomes more evident. For example, ConVAT  achieves as much as 3.4\% and 5.8\% improvement on accuracy in the uniform and random noise settings where the noise rate is set to highest.  When it comes to the comparison with VAT, ConVAT exhibits superior performance in almost every settings. These results clearly demonstrates that infusing perturbation into context-level is more effective than infusing it into input-level. That is to say, ConVAT has more resiliency on noisy labels, notably on extreme cases.





\section{Analysis}

In this section, we analyze the proposed method in depth. We first examine whether the proposed method could maintain accuracy without memorizing label noises in Section 5.1. We then describe the choice of the epsilon $\epsilon$ (in Eq. \ref{radius2}) at various noise levels in Section 5.2. Lastly, we confirm whether our method is indeed superior to the normal virtual adversarial training \cite{miayto2016virtual} in terms of time and memory efficiency.


\subsection{Denoising Effect of ConVAT}

\begin{figure*}[hbt!]
\centering     
\subfloat[]{\includegraphics[width=50mm]{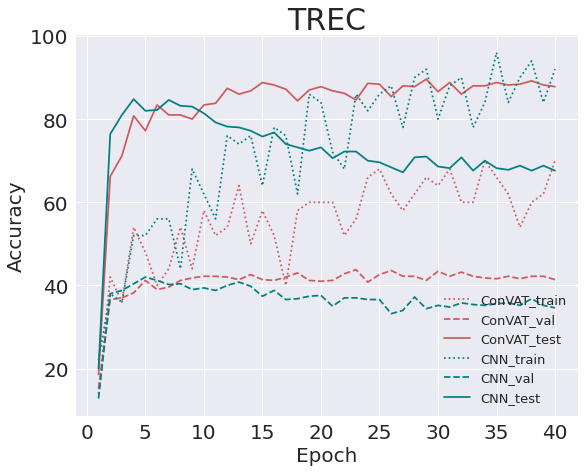}}
\subfloat[]{\includegraphics[width=50mm]{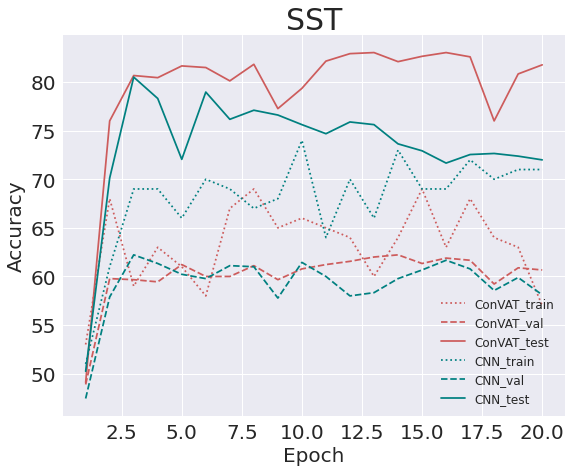}}
\subfloat[]{\includegraphics[width=50mm]{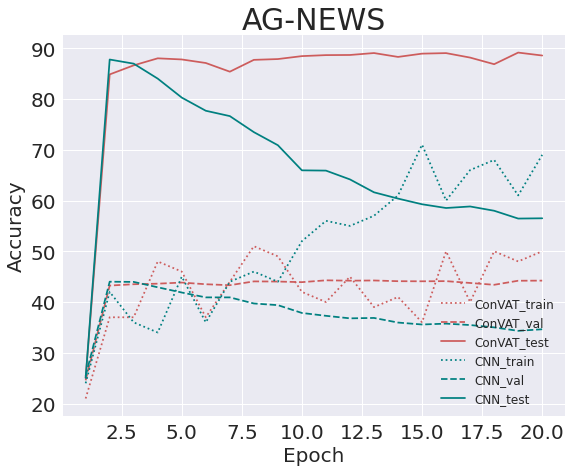}}
\caption{Train, validation and test accuracies versus training epochs on each datasets.}
\label{deno}
\end{figure*}

To analyze the denoising effects of ConVAT, we train ConVAT and CNN with the noisy datasets and recorded the accuracy of train, validation, and test accuracy for every epoch. We plot the results in Figure \ref{deno}. In this analysis, we train each method on TREC, SST, and Ag-News where 50\% of training samples are corrupted by uniform noises. Not surprisingly, we observe that train accuracies of CNN tend to keep increasing over training epochs and, however, the test accuracies drop rapidly after certain points where the memorization occurs. This performance trend can be widely observed in previous literature \cite{malach2017decoupling,han2018co}. However, ConVAT not only maintains its performance robustly, but it is also increasing little by little. This result indicates that the classifier with ConVAT does not overfit to noisy labels. This effect can be observed in the entire datasets. 

\begin{figure*}[hbt!]
\centering     
\subfloat[]{\includegraphics[width=50mm]{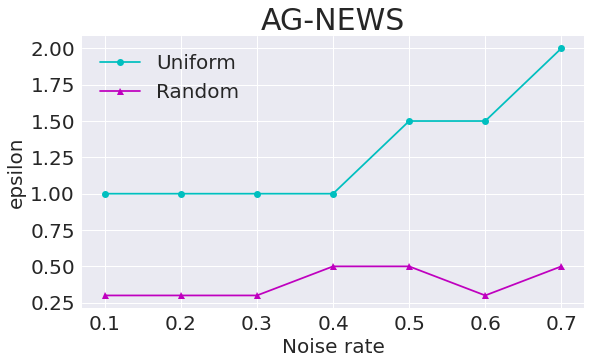}}
\subfloat[]{\includegraphics[width=50mm]{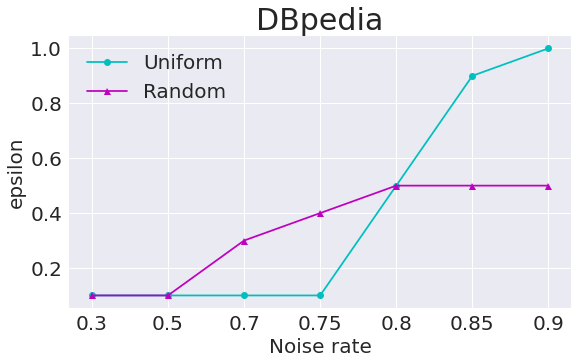}}
\subfloat[]{\includegraphics[width=50mm]{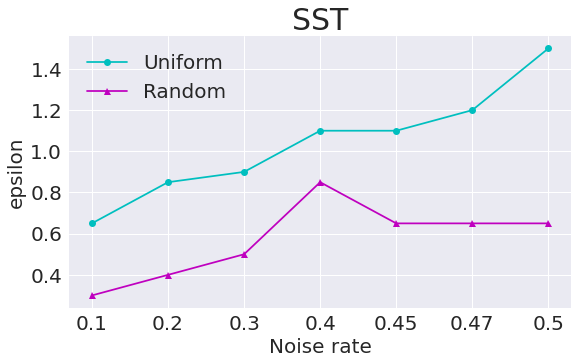}}
\caption{Optimal value of epsilon $\epsilon$ on the validation performance for
supervised task on AG-NEWS, DBpedia, SST}
\label{eps}
\end{figure*}

\begin{figure*}[hbt!]

\centering
\includegraphics[width=\textwidth]{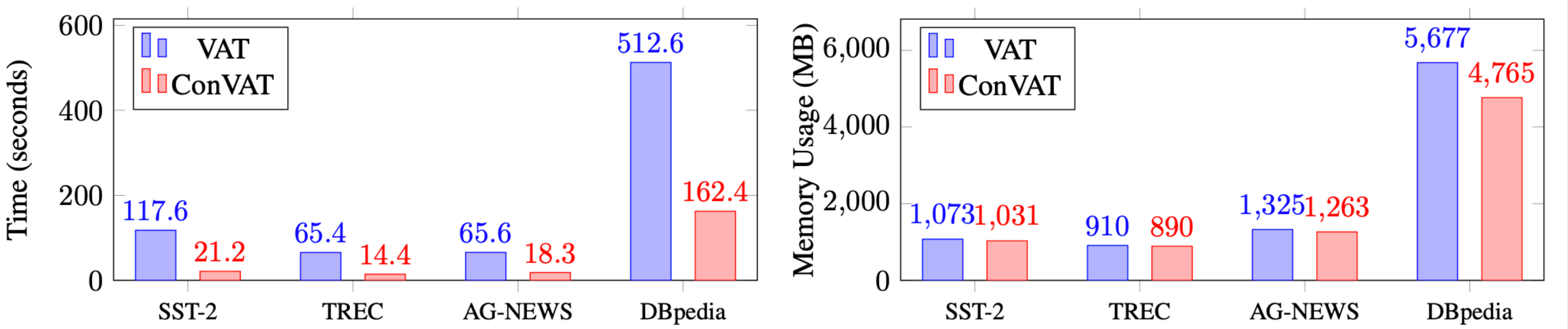}
    \caption{Time and memory usage comparison between ConVAT and VAT.}
    \label{time-memo}
\end{figure*}

\begin{figure*}[hbt!]
\centering     
\subfloat[Context vectors from ConVAT]{\includegraphics[width=130mm]{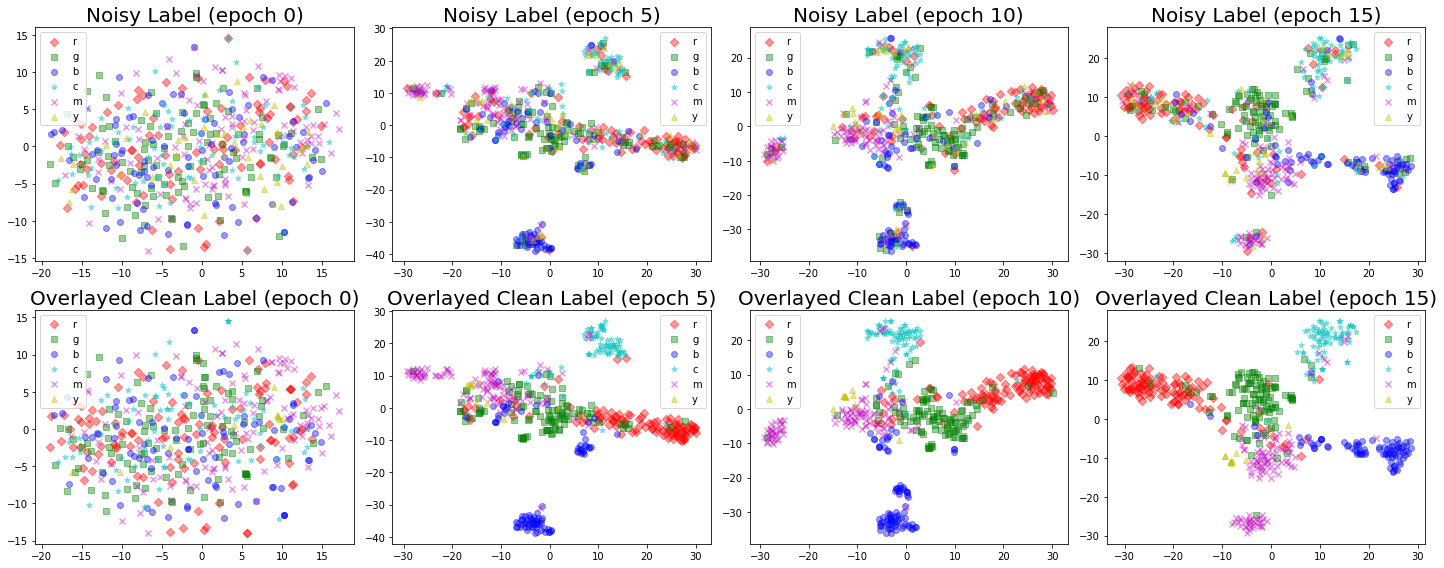}}

\subfloat[Context vectors from CNN]{\includegraphics[width=130mm]{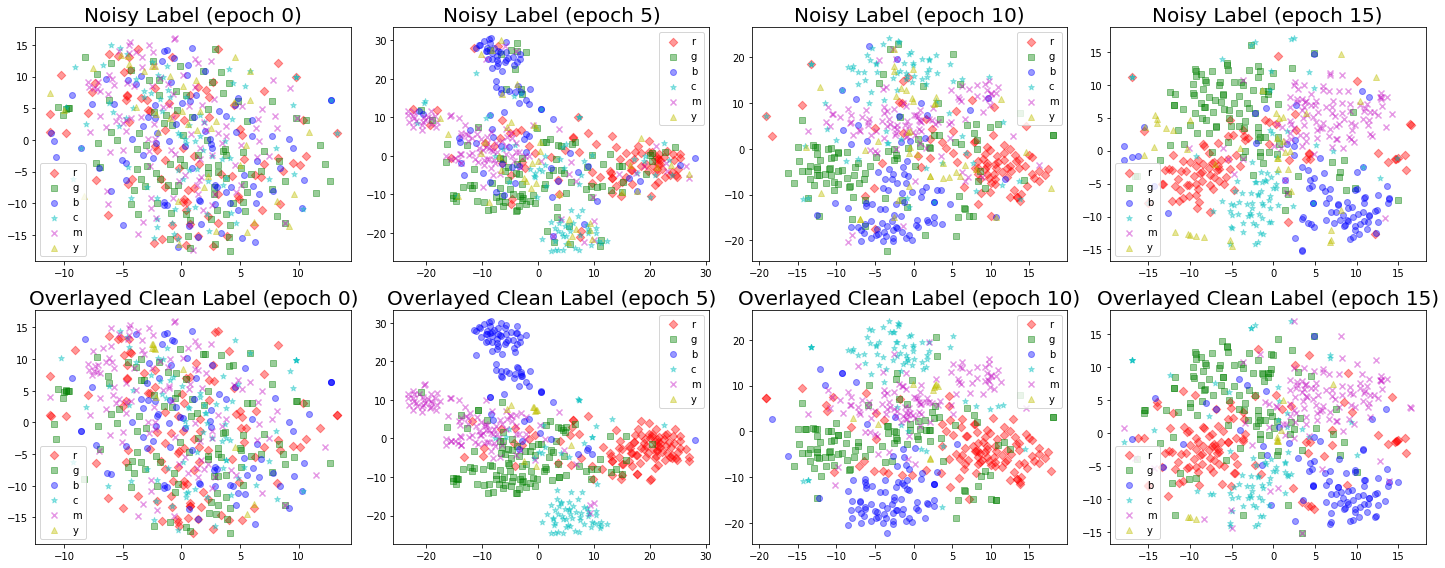}}
\caption{tSNE visualization of the context vectors from ConVAT and CNN.}\label{vis}
\end{figure*}

To further analyze the effect, we visualize the context vectors of each sample from TREC datasets with 50\% uniform noises through t-SNE \cite{van2014accelerating} and this is shown in Figure \ref{vis}. The context vectors from ConVAT is represented in Figure \ref{vis} (a) and vectors from CNN is visualized in Figure \ref{vis} (b). We first found that vectors from ConVAT stay clustered and are arranged more distinctively as the training proceeds. We believe that such well-clustered results come from the smoothing effect between context vectors and near data points and this smoothing prevents the model from overfitting to noisy labels. In contrast, the vectors from CNN shows some clustered result on early training epochs, but it starts to lose each cluster as training proceeds. This shows that CNN overfits noisy labels and loses the ability to discriminate each data points into their classes.

\subsection{Choice of Perturbation length $\epsilon$}

Epsilon in Eq. \ref{radius2} plays a role in determining how far a data point considers the neighbors. If the epsilon is small, the data point only takes into account the close distance neighbor which has similar logits. If the epsilon is big, data points see more diverse contexts that lead the model to be more general and labels to be smoother. Therefore the optimal result of the dataset with a low noise rate can be driven with small epsilon. On the other hand, when training with a higher noise rate, the model must consider a longer range of neighbors. 

We confirm whether such epsilon choice is indeed effective by finding optimal epsilon values through a grid search. The search space ranges from 0.0 to 3.0 with 0.1 steps. The best performing epsilon on datasets is shown in Figure \ref{eps}. As can be seen from all datasets, the optimal epsilon increases when the noise rate in datasets is large. This result supports our strategy of choosing proper epsilon. To find the best epsilon on datasets in practice, estimating the noise rate is required. Fortunately, noise rate can be easily estimated in practice \cite{liu2015classification,patrini2017making} and we, therefore, could find best performing hyper-parameters without an exhaustive grid search.



\subsection{Time and Memory usage Analysis of ConVAT}

We design the proposed method to handle label noises instead of adversarial inputs. We thus have strong advantages in terms of both time and memory efficiency compared to the normal VAT \cite{miayto2016virtual}. Specifically, previous works perform a forward and backward propagation twice to calculate a perturbation, but ConVAT propagates the whole network once. It results in that the computational cost of ConVAT does not depend on network architecture unlike VAT. We quantitatively analyze the effect on entire datasets and Figure \ref{time-memo} shows the comparison results. As can be seen from the figure (left), ConVAT requires 5x less time cost compared to VAT. The computational time gap becomes bigger as the size of the networks is larger. 

We also have found that our method requires fewer memory footprints than the VAT. The difference comes from the process of calculating worst-case perturbations (Eq. \ref{radius2}). Specifically, our method only requires the gradients of the softmax layer while VAT requires every gradient of entire networks (i.e., embedding layer to softmax layer).




\section{Conclusion}
In this paper, we have proposed context-based virtual adversarial training, coined ConVAT, that is a robust training method against label noises. Unlike previous works, ConVAT is designed as a network-agnostic manner and does not require additional training parameters. Comprehensive evaluation results have clearly shown that ConVAT is superior to previous works and has strong advantages in terms of the time complexity. Furthermore, we analyze our method in-depth and found that ConVAT robustly prevents networks from overfitting to label noises. As future work, we plan to apply our method to different natural language tasks such as a sequence labeling task where noise labels frequently occur.

\section{References}\label{reference}

\bibliographystyle{lrec2022-bib}
\bibliography{lrec2022-example}


\end{document}